\def\year{2019}\relax
\documentclass[letterpaper]{article} 
\usepackage{aaai19}  
\usepackage{times}  
\usepackage{helvet}  
\usepackage{courier}  
\usepackage{url}  
\usepackage{graphicx}  

\usepackage{bm}
\usepackage{amssymb}
\usepackage{graphicx}
\usepackage{subfigure}
\usepackage{mathrsfs}
\usepackage{amsmath}
\usepackage{url}
\usepackage{algorithm}
\usepackage{algorithmic}
\usepackage{color}
\usepackage{tikz}
\usepackage{booktabs}
\usepackage{comment}
\usepackage{subeqnarray}
\usepackage{wrapfig}

\newcommand{\ignore}[1]{}

\newcommand{\ds}{SVM-DSN}

\begin{document}
%

\title{SVM-based Deep Stacking Networks}

\author{Jingyuan Wang$^{\dag, \S}$, Kai Feng$^{\dag}$, Junjie Wu$^{\ddag, \S}$\thanks{Corresponding author}\\
$\dag$ MOE Engineering Research Center of Advanced Computer Application Technology, \\ School of Computer Science Engineering, Beihang University, Beijing 100191, China\\
$\ddag$ Beijing Key Laboratory of Emergency Support Simulation Technologies for City Operations, \\ School of Economics and Management, Beihang University, Beijing 100191, China\\
$\S$ Beijing Advanced Innovation Center for Big Data and Brain Computing, \\ Beihang University, Beijing 100191, China\\
Email: \{jywang, fengkai, wujj\}@buaa.edu.cn
}

\maketitle

\begin{abstract}
The deep network model, with the majority built on neural networks, has been proved to be a powerful framework to represent complex data for high performance machine learning. In recent years, more and more studies turn to non-neural network approaches to build diverse deep structures, and the Deep Stacking Network (DSN) model is one of such approaches that uses stacked easy-to-learn blocks to build a parameter-training-parallelizable deep network. In this paper, we propose a novel SVM-based Deep Stacking Network (\ds), which uses the DSN architecture to organize linear SVM classifiers for deep learning. A BP-like layer tuning scheme is also proposed to ensure holistic and local optimizations of stacked SVMs simultaneously. Some good math properties of SVM, such as the convex optimization, is introduced into the DSN framework by our model. From a global view, \ds~can iteratively extract data representations layer by layer as a deep neural network but with parallelizability, and from a local view, each stacked SVM can converge to its optimal solution and obtain the support vectors, which compared with neural networks could lead to interesting improvements in anti-saturation and interpretability. Experimental results on both image and text data sets demonstrate the excellent performances of \ds~compared with some competitive benchmark models.
\end{abstract}

\section{Introduction}
Recent years have witnessed the tremendous interests from both the academy and industries in building deep neural networks~\cite{deeplearning,representationlearning}. Many types of deep neural networks have been proposed for classification, regression and feature extracting tasks, such as Stacked Denoising Autoencoders (SAE)~\cite{SAE}, Deep Belief Networks (DBN)~\cite{dbn}, deep Convolutional Neural Networks (CNN)~\cite{cnn}, Recurrent Neural Networks (RNN)~\cite{rnn}, and so on.

Meanwhile, the shortcomings of neural network based deep models, such as the non-convex optimization, hard-to-parallelizing, and lacking model interpretation, are getting more and more attentions from the pertinent research societies. Some potential solutions have been proposed to build deep structure models using non neural network approaches. For instance, in the literature, the PCANet build a deep model using an unsupervised convolutional principal component analysis~\cite{pcanet}. The gcForest builds a tree based deep model using stacked random forests, which is regarded as a good alternative to deep neural networks~\cite{ijcai2017-497}. Deep Fisher Networks build deep networks by stacking Fisher vector encoding into multiple layers~\cite{dfn}.

Along this line, in this paper, we propose a novel SVM-based Deep Stacking Network (\ds) for deep machine learning.  On one hand, \ds~belongs to the community of Deep Stacking Networks (DSN), which consist of many stacked multilayer base blocks that could be trained in a parallel way and have comparable performance with deep neural networks~\cite{dcn,dsn}. In this way, \ds~can gain the deep learning ability with extra scalability. On the other hand, we replace the traditional base blocks in a DSN, {\it i.e.}, the perceptrons, by the well known Support Vector Machine (SVM), which has long been regarded as a succinct model with appealing math properties such as the convexity in optimization, and was considered as a different method to model complicated data distributions compared with deep neural networks~\cite{bengio2009learning}. In this way, \ds~can gain the ability in anti-saturation and enjoys improved interpretability, which are deemed to be the tough challenges to deep neural networks. A BP-like Layered Tuning (BLT) algorithm is then proposed for \ds~to conduct holistic and local optimizations for all base SVMs simultaneously.

Compared with the traditional deep stacking networks and deep neural networks, the \ds~model has the following advantages:
\begin{itemize}
\item The optimization of each base-SVM is convex. Using the proposed BLT algorithm, all base-SVMs are optimized as a whole, and meanwhile each base-SVM can also converge to its own optimum. The final solution of \ds~is a group of optimized linear SVMs that are integrated as a deep model. This advantage allows \ds~to avoid the neuron saturation problem in deep neural networks, and thus could improve the performance.

\item The \ds~model is very easy to parallelize. The training parallelization in \ds~can reach the base-SVM level due to the support vectors oriented property of SVM, but the traditional DSN can only reach the block level.

\item The \ds~model has improved interpretability. The support vectors in base-SVMs can provide some insightful information about what a block learned from training data. This property empowers users to partially understand the feature extracting process of the \ds~model.
\end{itemize}

\begin{figure}[t!]
    \centering
    \includegraphics[width=5cm]{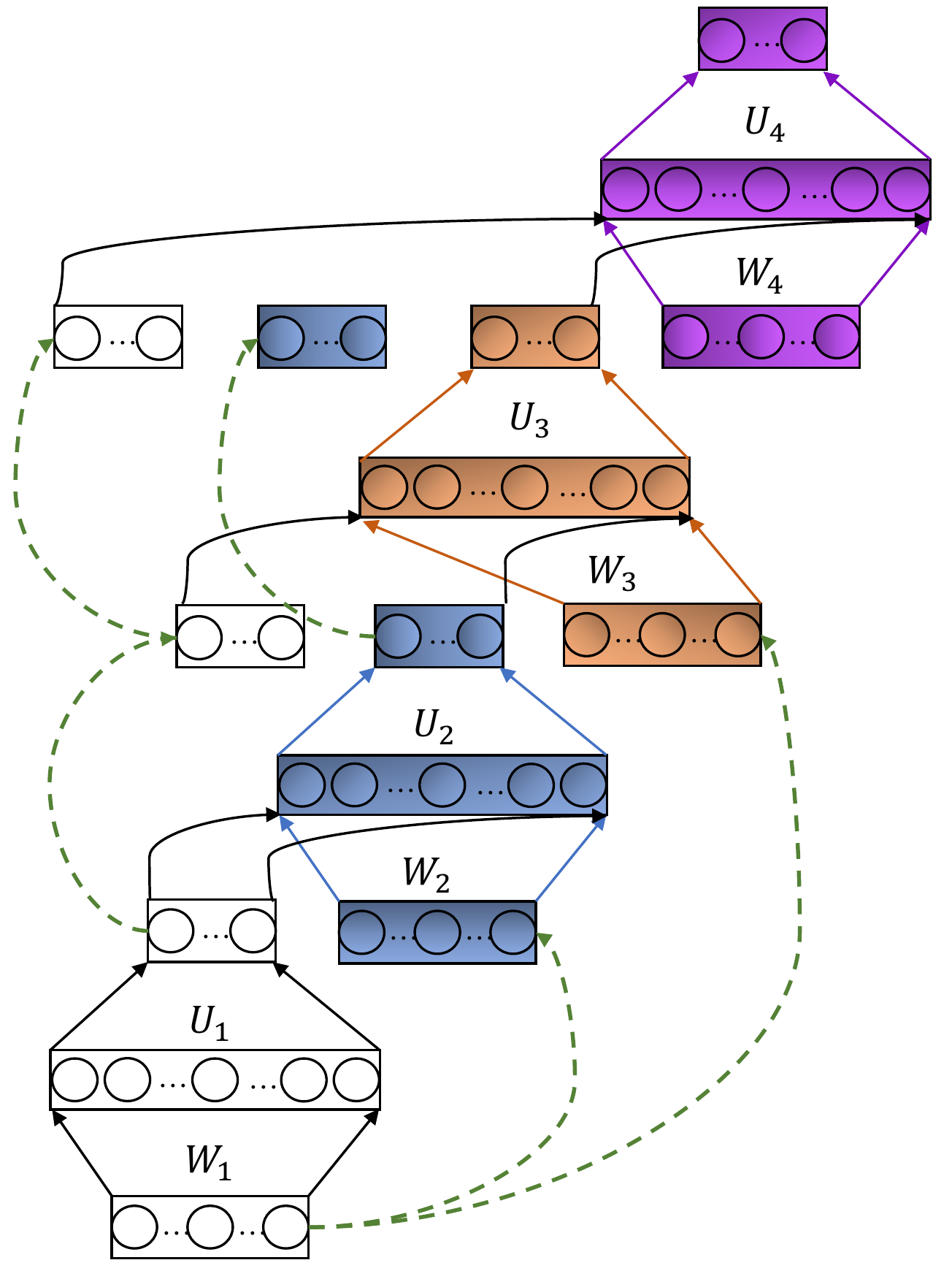}
    \caption{An illustration of the DSN architecture~\cite{dsn}. The color is used to distinguish different blocks in a DSN. The components in the same color belong to the same block.}\label{fig:dsn_architecture}
\end{figure}

Experimental results on image and sentiment classification tasks show that \ds~model obtains respectable improvements over neural networks. Moreover, compared with the stacking models with strong base-learners, the \ds~model also demonstrates significant advantages in performance.

\section{The SVM-DSN Model}

\subsection{Framework of Deep Stacking Network}

The Deep Stacking Network is a scalable deep machine learning architecture~\cite{dsn,dcn} that consists of stacked easy-to-learn blocks in a layer by layer manner. In the standard DSN, a block is a simplified multilayer perceptron with a single hidden layer. Let the inputs of a block be a vector $\bm{x}$, the block uses a connection weight matrix $\bm{W}$ to calculate the hidden layer vector $\bm{h}$ as
\begin{equation}\label{}
  \bm{h} = \varphi \left( \bm{W}^\top \bm{x} \right),
\end{equation}
where $\varphi(x) = 1/(1+\exp(-x))$ is a sigmoid nonlinear activation function. Using a weight matrix $\bm{U}$, the objective function of the DSN block optimization is defined as
\begin{equation}\label{eq:convex}
  \min \left\| \bm{y} - \bm{U}^\top \bm{h} \right\|_F^2.
\end{equation}

As shown in Fig.~\ref{fig:dsn_architecture}, the blocks of a DSN are stacked layer by layer. For the block in the input layer, the input vector $\bm{x}$ contains only the raw input features. For blocks in the middle layers, $\bm{x}$ is a concatenated vector of the raw input features and output representations of all previous layer blocks.

The training of deep stacking networks contains two steps: block training and fine-tuning. In the block training step, the DSN blocks are independently training as supervised multilayer perceptrons. In the fine-tuning step, all the stacked blocks are considered as a multi-layer deep neural network. The parameters of DSN are end-to-end trained using the error Back Propagation (BP) algorithm.

\subsection{SVM-based DSN Blocks}

In the SVM-DSN model, we adopt support vector machines to implement a DSN block. A SVM classifier is a hyperplane $\bm{\omega}^{\top} \bm{x} + b = 0$ that divides the feature space of a data sample $\bm{x}$ into two parts --- one for the positive and the other for the negative. The parameters $\bm{\omega}$ and $b$ are optimized to maximize the minimum distances from the hyperplane to a set of training samples $T = \{(\bm{x}_k,y_k)| y_k \in \{-1,1\}, k = 1, \ldots, K\}$, {\it i.e.},
\begin{equation}~\label{equ:svm}
    \begin{aligned}
    \max_{\bm{\omega}, b} & ~~\quad \frac{2}{\|\bm{\omega}\|} \\
    s.t. & ~~\quad y_k (\bm{\omega}^\top \bm{x}_k + b) \geq 1, \quad k = 1,2, \ldots, N.
    \end{aligned}
\end{equation}
A training sample is called a \emph{support vector} if the constraint in Eq.~\eqref{equ:svm} turns into equality.

For a multi-class problem with $N$ classes, we connect the input vector $\bm{x}$ of a DSN block with $N$ binary SVM classifiers --- each for recognizing whether a sample belongs to a corresponding class --- to predict the label of a sample. A binary SVM classifier in a DSN block is called a {\em base-SVM}. The $N$ binary SVM classifiers for a $N$ classification problem is called as a {\em base-SVM group}. A \ds~block could contains multiple base-SVM groups. In the same block, all base-SVM groups share the same input vector $\bm{x}$.

\subsection{Stacking Blocks}

Given a classification hyperplane of a SVM, the decision function for the sample $\bm{x}_k$ is expressed as
\begin{equation}\label{}
  f(\bm{x}_k) = \mathrm{sign}\left(\bm{\omega}^\top \bm{x}_k + b\right),
\end{equation}
where $f(\bm{x}_k) = 1$ for the positive class and $f(\bm{x}_k) = -1$ for the negative. The distance from a sample to the hyperplane could be considered as the \emph{confidence} of a classification decision. For the samples behind the support vectors, {\it i.e.}, $\left|\bm{\omega}^{\top} \bm{x}_k + b \right| > 1$, the confidence is 1, otherwise is $\left|\bm{\omega}^{\top} \bm{x}_k + b \right|$. We therefore can express the classification confidence of a SVM classifier for the sample $\bm{x}_k$ as
\begin{equation}\label{}
  g(\bm{x}_k) = \min\left(1, |\bm{\omega}^{\top} \bm{x}_k + b | \right).
\end{equation}

We denote the $i$-th base-SVM in the layer $l$ as $svm(l,i)$ and its decision function and confidence as $f^{(l,i)}(\cdot)$ and $g^{(l,i)}(\cdot)$, respectively. For the base-SVM $svm(l,i)$, we define a \emph{confidence weighted output} $y^{(l,i)}$ as
\begin{equation}\label{eq:svm_output}
  y^{(l,i)} = f^{(l,i)}(\bm{x})\cdot g^{(l,i)}(\bm{x}).
\end{equation}
In the layer $l$+1, \ds~concatenates the confidence weighted outputs of all base-SVMs in the previous layers and raw inputs as
\begin{equation}\label{eq:virtualx}
\begin{aligned}
  \bm{x}^{(l+1)} = &\left(y^{(l,1)}, \ldots, y^{(l,i)}, \ldots,
   y^{(l-1,1)}, \ldots, y^{(l-1,i)}, \right. \\
  &\left. \ldots, y^{(1,1)}, \ldots, y^{(1,i)}, \ldots,
   x^{(1,1)} , \ldots, x^{(1,i)} \right)^\top.
\end{aligned}
\end{equation}
The base-SVMs in the layer $l$+1 use $\bm{x}^{(l+1)}$ as the input to generate their confidence weighted outputs $y^{(l+1,i)}$. In this way, base-SVMs are stacked and connected layer by layer.

\section{Model Training}

\subsection{Block Training}

Similar to the standard deep stacking network, the training of the \ds~model also contains a block training step and a fine-tuning step.

In the block training step, the base-SVMs in a DSN block are trained as regular SVM classifiers. Given a set of training samples $T = \{(\bm{x}_k,y_k)| k = 1, \ldots, K\}$, where $y_k$ is the ground-truth label of $\bm{x}_k$, the objective function of a base-SVM group with $N$ classification is defined as
\begin{equation}\label{eq:loss_svm}
\begin{aligned}
  \mathcal{J} & =  \frac{1}{2}\left\|\bm{\Omega}\right\|_F^{2} \\
  & + C \sum_{k=1}^K \sum_{i=1}^N \ell_{hinge} \left(y^{(i)}_k(\bm{\omega}^{(i)\top}{\bm x}_k + b^{(i)})\right),
\end{aligned}
\end{equation}
where $\bm{\Omega} = \left(\bm{\omega}^{(1)\top}, \ldots, \bm{\omega}^{(N)\top}\right)$, and $y^{(i)}_k=1$ if $y_k = i$ and -1 otherwise. The function $\ell_{hinge}(\cdot)$ is a hinge loss function defined as $\ell_{hinge}(z) = \max\left(0, 1 - z\right)$. The parameter $\theta = \left\{(\bm \omega^{(i)}, b^{(i)})|\forall~i \right\}$ is inferred as ${\theta} = \arg\min  \limits_{{\theta}} \mathcal{J}(\theta)$.

In order to increase the diversity of base-SVM groups in a block, we adopt a bootstrap aggregating method in the block training. For a block with $M$ base-SVM groups, we re-sample the training data as $M$ sets using the bootstrap method~\cite{bootstrap}. Each base-SVM group is trained using one re-sampled data set.

\subsection{Fine Tuning}

The traditional DSN model is based on neural networks and uses the BP algorithm in the fine-tuning step. For the \ds~model, we introduce SVM training into the BP algorithm framework, and propose a BP-like Layered Tuning (BLT) algorithm to fine-tune the model parameters.

Algorithm~\ref{alg:SLT} gives the pseudocodes of BLT. In general, BLT iteratively optimizes the base-SVMs from the output layer to the input layer. In each iteration, BLT optimizes $svm(l,i)$ by firstly generating a set of \emph{virtual} training samples $T^{(l,i)} = \{(\bm{x}^{(l)}_k,\tilde{y}^{(l,i)}_k)| k = 1, \ldots, K\}$, and then trains a new $svm(l,i)$ on $T^{(l,i)}$.

According to Eq.~\eqref{eq:svm_output} and Eq.~\eqref{eq:virtualx}, it is easy to have $\bm{x}^{(l)}_k = (y^{(l-1,1)}_k,y^{(l-1,2)}_k,\cdots,y^{(l-1,i)}_k,\cdots)^{\top}$. However, the calculation of the \emph{virtual label} $\tilde{y}^{(l,i)}_k$ is not that straightforward. Specifically, BLT adopts a gradient descent method to calculate $\tilde{y}^{(l,i)}_k$ as
\begin{equation}\label{eq:virtual_y}
  \tilde{y}^{(l,i)}_k = \sigma\left({{y}}^{(l,i)}_k - \eta \left.\frac{\partial \mathcal{J}^{(o)}}{\partial {{y}}^{(l,i)}}\right|_{y^{(l,i)}=y^{(l,i)}_k}\right),
\end{equation}
where $\mathcal{J}^{(o)}$ is the objective function of the output layer, $y^{(l,i)}_k$ is the output of $\bm{x}^{(l)}_k$ in the previous iteration, $\eta$ is the learning rate, and $\sigma(\cdot)$ is a shaping function defined as
\begin{equation}~\label{equ:svm_act}
\sigma(z) = \left\{
    \begin{aligned}
    1, & \quad  ~z~ > 1 \\
    z, & \quad  |z| \leq 1 \\
    -1, & \quad ~~ < -1
    \end{aligned}.
\right.
\end{equation}
Note that since the term $- \eta {\partial \mathcal{J}^{(o)}}/{\partial {{y}}^{(l,i)}}$ in Eq.~\eqref{eq:virtual_y} is a negative gradient direction of $\mathcal{J}^{(o)}$, tuning the output ${y}^{(l,i)}$ to the virtual label $\tilde{{y}}^{(l,i)}$ can reduce the value of the objective function $\mathcal{J}^{(o)}$ in the output layer. Therefore, it could be expected that BLT can lower the overall model prediction error iteratively by training base-SVMs on virtual training sets in each iteration.

\begin{algorithm}[t]\small
	\caption{BP-like Layered Tuning Algorithm}
	\begin{algorithmic}[1]\label{alg:SLT}
		\STATE {\em Initialization}: Initializing $\bm{\omega}^{(l,i)}, b^{(l,i)}$ for all $svm(l,i)$ as random values.
		\REPEAT
        \STATE Select a batch of training samples $T = \{(\bm{x}_k,{{y}}_k)| k = 1, \ldots, K\}$.
        \FOR {$l=L, L-1, \ldots, 2, 1$}
        \FOR {$i=1, 2, \ldots$}
        \STATE  Use Eq.~\eqref{eq:svm_output}, Eq.~\eqref{eq:virtualx}, and Eq.~\eqref{eq:virtual_y} to calculate $T^{(l,i)} = \{(\bm{x}^{(l)}_k,\tilde{y}^{(l,i)}_k)| k = 1, \ldots, K\}$.
        \STATE Use $T^{(l,i)}$ to train $svm(l,i)$ as Eq.~\eqref{eq:obj_subsvm}.
        \ENDFOR
        \ENDFOR
		\UNTIL The algorithm converges.
	\end{algorithmic}
\end{algorithm}

Given the training set $T^{(l,i)} = \{(\bm{x}_k^{(l)}, \tilde{{y}}_k^{(l,i)})|k = 1, \ldots K\}$, the objective function of training $svm(l,i)$ is defined as
\begin{equation}\label{eq:obj_subsvm}\small 
\begin{aligned}
\min \mathcal{J}^{(l,i)} & =  \frac{1}{2} \left\|\bm{\omega}^{(l,i)}\right\|^{2} \\
& + \underbrace{C_{1}\sum_{k \in \Theta} \ell_{hinge}\left(\tilde{y}^{(l,i)}_k(\bm{\omega}^{(l,i)\top}{\bm{x}}^{(l)}_k + b^{(l,i)})\right)}_{\mathrm{The~SVM~Loss}} \\
& + \underbrace{C_{2}\sum_{k \notin \Theta}\ell_{\epsilon}\left(\bm{\omega}^{(l,i)\top}{\bm{x}}^{(l)}_k + b^{(l,i)} - \tilde{y}^{(l,i)}_k \right)}_{\mathrm{The~SVR~Loss}},
\end{aligned}
\end{equation}
where $\Theta$ is the index set of the virtual labels $\left| \tilde{y}_k^{(l,i)} \right| = 1$, and the function $\ell_{\epsilon}(\cdot)$ is an $\epsilon$-insensitive loss function in the form of $\ell_{\epsilon}(z)=\max(|z|-\epsilon,0)$.

Note that the objective function in Eq.~\eqref{eq:obj_subsvm} contains two types of loss functions so as to adapt to the different conditions of $\tilde{y}^{(l,i)}_k$. When $\tilde{y}^{(l,i)}_k \in \{-1,1\}$, {\it i.e.}, the virtual labels are binary, BLT trains $svm(l,i)$ as the standard SVM classifier and thus uses the hinge loss function to measure errors. When $\tilde{y}^{(l,i)}_k \in (-1,1)$, the objective function adopts a Support Vector Regression loss term $\ell_{\epsilon}$ for this condition. In the Appendix, we prove that the problem defined in Eq.~\eqref{eq:obj_subsvm} is a quadratic convex optimization problem. The training of $svm(l,i)$ can thus reach an optimal solution by using various quadratic programming methods such as sequential minimal optimization and gradient descents.

We finally turn to the small problem unsolved --- how to calculate the partial derivative ${\partial \mathcal{J}}^{(o)}/{\partial {y}^{(l,i)}}$ in Eq.~\eqref{eq:virtual_y}. Based on the chain rule, the partial derivative can be recursively calculated as
\begin{equation}\label{eq:partial_2}
\begin{aligned}
\frac{\partial \mathcal{J}}{\partial {y}^{(l,i)}} & =  \sum_{m=l+1}^L \sum_j \frac{\partial \mathcal{J}}{\partial {y}^{(m,j)}}\frac{\mathrm{d} {y}^{(m,j)}}{\mathrm{d} {z}^{(m,j)}}\frac{\partial {z}^{(m,j)}}{\partial {y}^{(l,i)}} \\
& = \sum_{m=l+1}^L \sum_j \frac{\partial \mathcal{J}}{\partial {y}^{(m,j)}} y'\left({z}^{(m,j)}\right) \omega^{(m,{j})}_i,
\end{aligned}
\end{equation}
where $\omega^{(m,{j})}_i$ is the connection weight of ${y}^{(m,j)}$ in $svm(m,j)$, and ${z}^{(m,j)} = \bm{\omega}^{(m,j)\top} \bm{x}^{(m-1)} + b^{(m,j)}$. The term $y'(z)$ is the derivative of the function in Eq.~\eqref{eq:svm_output}, which is in the form of
\begin{equation}~\label{equ:d_act}
y'(z) = \left\{
    \begin{aligned}
    0, & \quad  |z| > 1 \\
    1, & \quad  |z| \leq 1
    \end{aligned}.
\right.
\end{equation}
The principle of this chain derivation is similar to the error back-propagation of the neural network training. The difference lies in that the BP algorithm calculates the derivative for each neuron connecting weight but BLT for each base-SVM output. That is why we name our algorithm as {\em BP-like} Layered Tuning.

\section{Model Properties}

\subsection{Connection to Neural Networks}~\label{sec:connect}
The \ds~model has close relations with neural networks. If we view the base-SVM output function defined in~Eq.~\eqref{eq:svm_output} as a neuron, the \ds~model can be regarded as a type of neural networks. Specifically, we can rewrite the function in Eq.~\eqref{eq:svm_output} as a neuron form as follows:
\begin{equation}\label{equ:svm_linear}
y^{(l,i)} = \sigma\left(\bm{\omega}^{^{(l,i)}\top} \bm{x}^{(l)} + b^{(l,i)}\right),
\end{equation}
where the shaping function $\sigma(\cdot)$ works as an activate function, with the output $\sigma(z)\in\{1,-1\}$ if $|z| \geq 1$, and $\sigma(z)=z$ if $|z| < 1$. As proved in~\cite{hornik1991approximation}, a multi-layer feedforward neural network with arbitrary bounded and non-constant activation function has an universal approximation capability. As a consequence, we could expect that the proposed \ds~model also has the universal approximation capability in theory.

Nevertheless, the difference between the \ds~model and neural networks is still  significant. Indeed, we have proven in the Appendix that the base-SVMs in our \ds~model have the following property: Given a set of {virtual} training samples $\{(\bm{x}^{(l)}_k,\tilde{y}^{(l,i)}_k)| k = 1, \ldots, K\}$ for $svm(l,i)$, to minimize the loss function defined in Eq.~\eqref{eq:obj_subsvm} is a convex optimization problem. Moreover, because the base-SVMs in the same block are mutually independent, the optimization of the whole block is a convex problem too. This implies that, in each iteration of the BLT algorithm, all blocks can converge to an optimal solution. In other words, \ds~ensures that all blocks and their base-SVMs ``do their own best'' to minimize their own objective functions in each iteration, which however is not the case for neural networks and MLP based deep stacking networks. It is also worth noting that this ``do their own best'' property is compatible with the decrease of the overall prediction error measured by the global objective function $\mathcal{J}^{(o)}$.

An important advantage empowered by the ``do their own best'' property is the {\em anti-saturation} feature of~\ds. In neural network models, the BP algorithm updates the parameter $\omega$ of a neuron as $\omega \leftarrow \eta {\partial \mathcal{J}}/{\partial \omega}$. Hence, the partial derivative for the $i$-th $\omega$ in the $j$-th neuron at the layer $l$ is calculated as
\begin{equation}\label{}
\begin{aligned}
\frac{\partial J}{\partial \omega^{(l,j)}_i} & =  \frac{\partial J}{\partial y^{(l,j)}} \frac{\mathrm{d} y^{(l,j)}}{\mathrm{d} z^{(l,j)}}  \frac{\partial z^{(l,j)}}{\partial \omega^{(l,j)}_i}  \\
& = \frac{\partial J}{\partial y^{(l,j)}}\cdot y'\left(z^{(l,j)}\right) \cdot y^{(l-1,i)},
\end{aligned}
\end{equation}
where $y'\left(z^{(l,j)}\right)$ is a derivative of the activation function. For the sigmoid activation function, if $|z|$ is very large then $y'$ becomes very small, and ${\partial J}/{\partial \omega} \rightarrow 0$. In this condition, the BP algorithm cannot update $\omega$ any more even if there is still much room for the optimization of $\omega$. This phenomenon is called the ``neuron saturation'' in neural network training. For the ReLU activation function, similar condition appears when $z<0$, where $y' = 0$ and ${\partial J}/{\partial \omega} = 0$. The neuron will die when a ReLU neuron fall into this condition.

In the BLT algorithm of \ds~model, the update of a base-SVM is guided by ${\partial \mathcal{J}}/{\partial {y}}$, with the details given in Eq.~\eqref{eq:partial_2}. From Eq.~\eqref{eq:partial_2}, we can see that unless all base-SVMs in an upper layer are saturated, {\it i.e.}, $y'({z}^{(m,j)})=0$ for all ${z}^{(m,j)}$, the base-SVMs in the layer $l$ would not fall into the saturation state. Therefore, we could expect that the saturation risk of a base-SVM in \ds~tends to be much lower than a neuron in neural networks.

\begin{figure*}[t]
	\centering
	\subfigure[Layer-1]{\label{fig:heat_h1} \includegraphics[width=0.452\columnwidth]{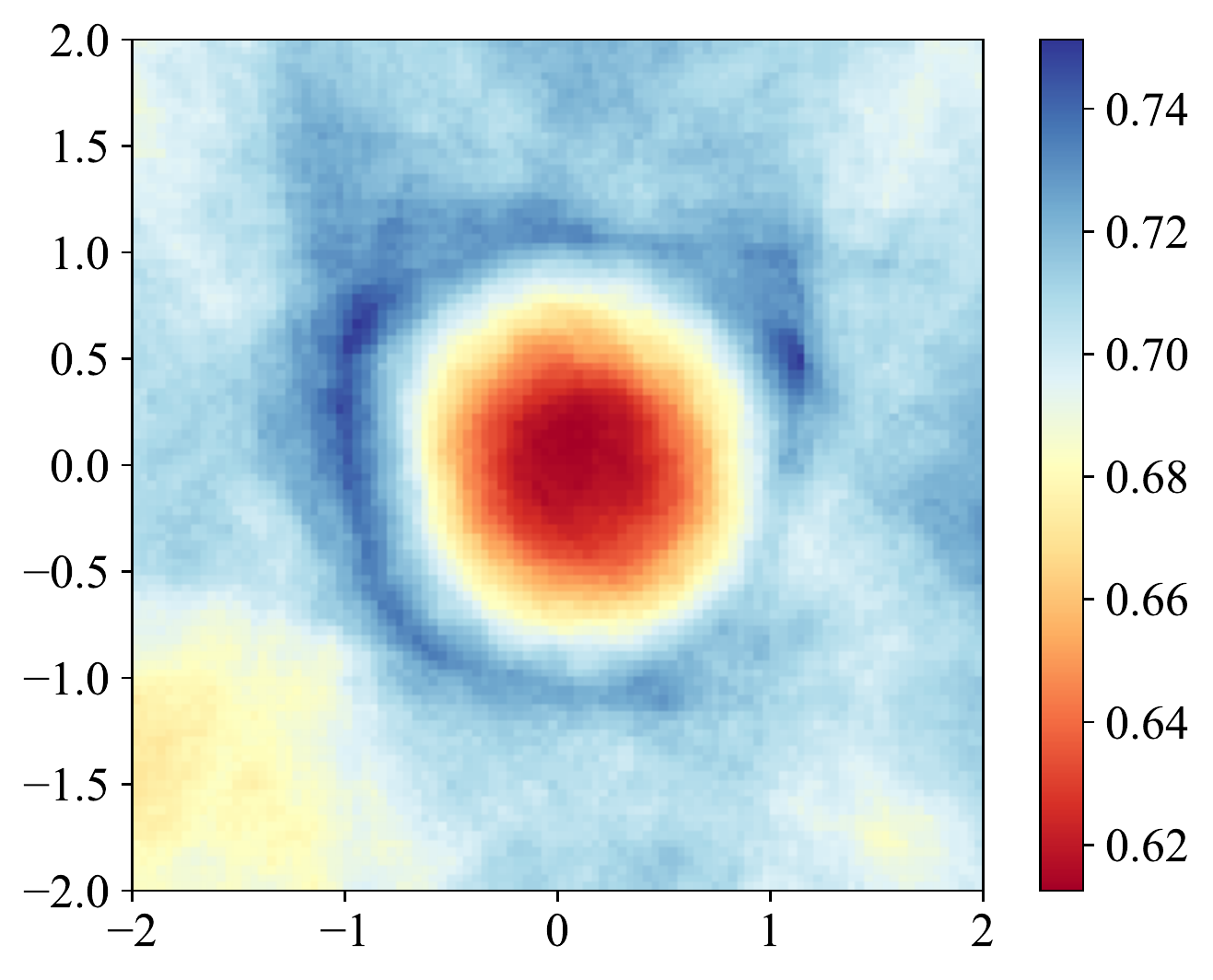}}~~~~~~~~~~
	\subfigure[Layer-2]{\label{fig:heat_h2}
\includegraphics[width=0.4532\columnwidth]{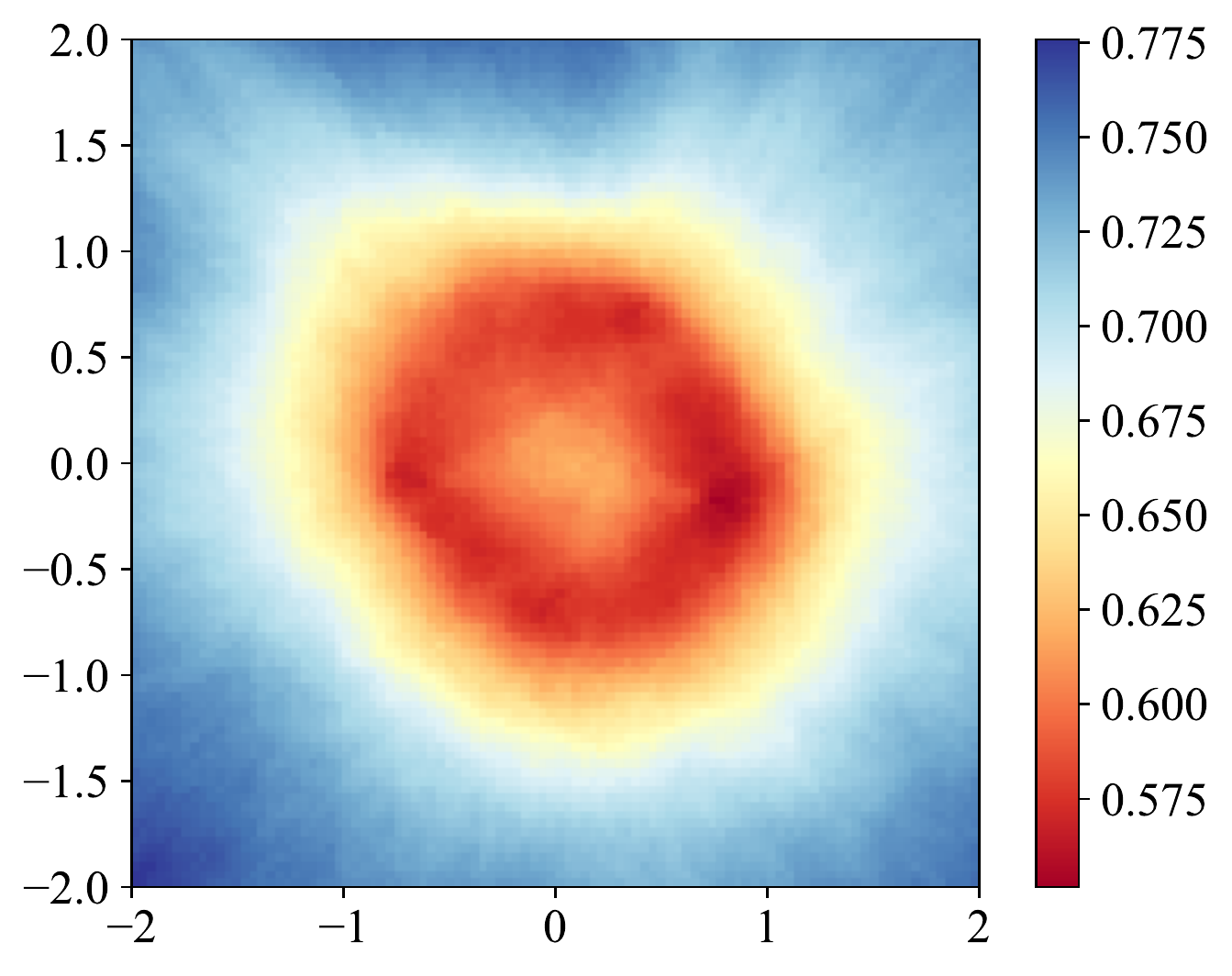}}~~~~~~~~
	\subfigure[Layer-3]{\label{fig:heat_h3} \includegraphics[width=0.4511\columnwidth]{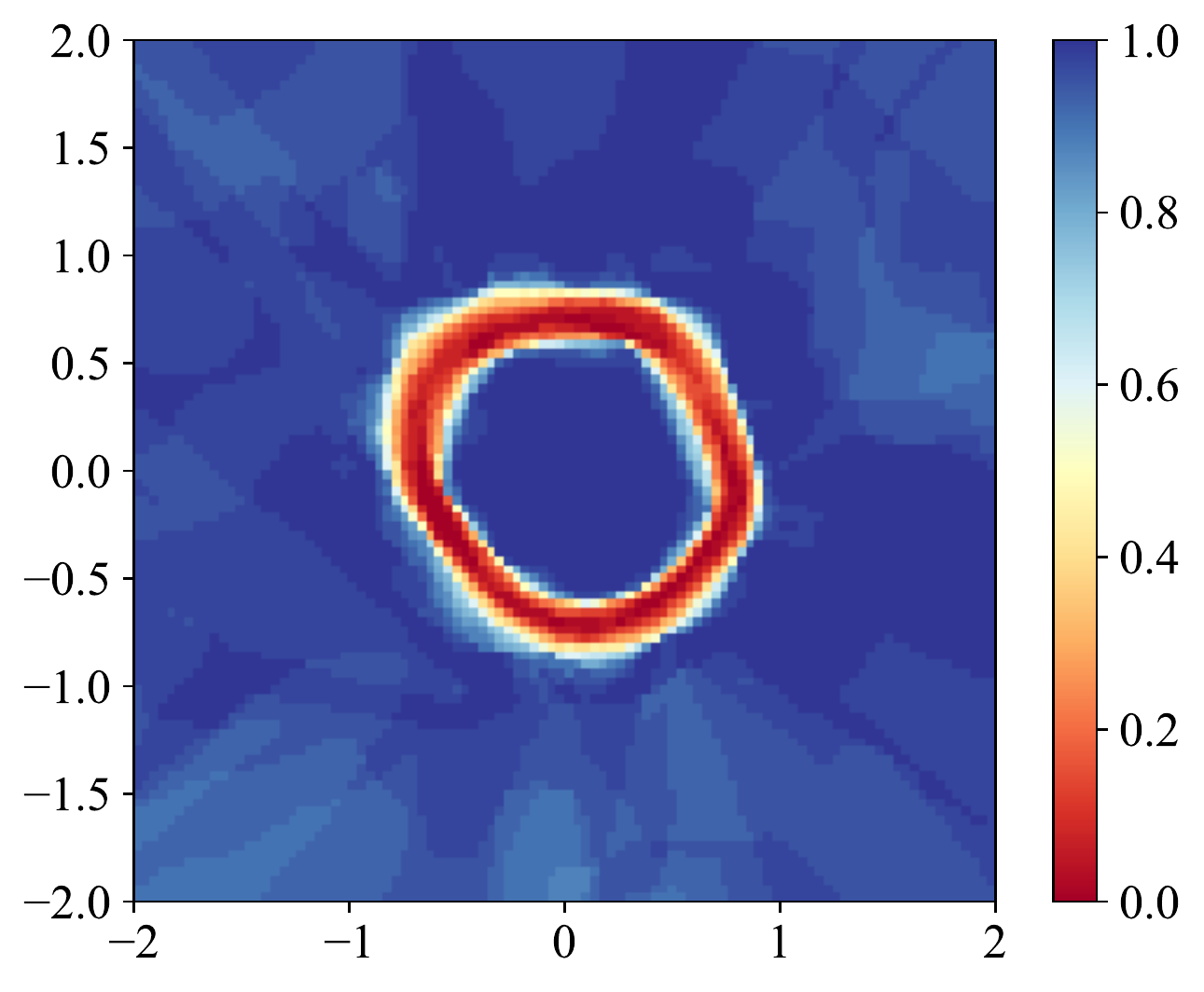}}
	\caption{The classifying plane maps of each layer in the \ds.}\label{fig:showcase}
\end{figure*}

\subsection{Interpretation}

In SVM classifiers, the support vector samples could provide some interpretation information about what a classifier learned from data set. The \ds~inherit this interpretation property of SVM. In the base-SVM output function defined in Eq.~\eqref{eq:svm_output}, the confidence function $g(\bm{x})$ indicates whether a sample is clearly classified by the hyperplane of a base-SVM. For a sample to be classified, we can calculate the average confidence of the sample in all base-SVM groups in a block. The average confidence indicates whether the feature extracted by the block and previous layers offer enough representations to identify label of the sample. Because the samples with low confidence are near to the hyperplane, we can use the low confidence samples to form a ``classifying plane'' map in each blocks. Comparing the ``classifying plane'' maps layer by layer, we could partly understand the feature extracting process of the stacked block in a \ds~model.

\begin{figure}[t]
    \centering
    \includegraphics[width=5cm]{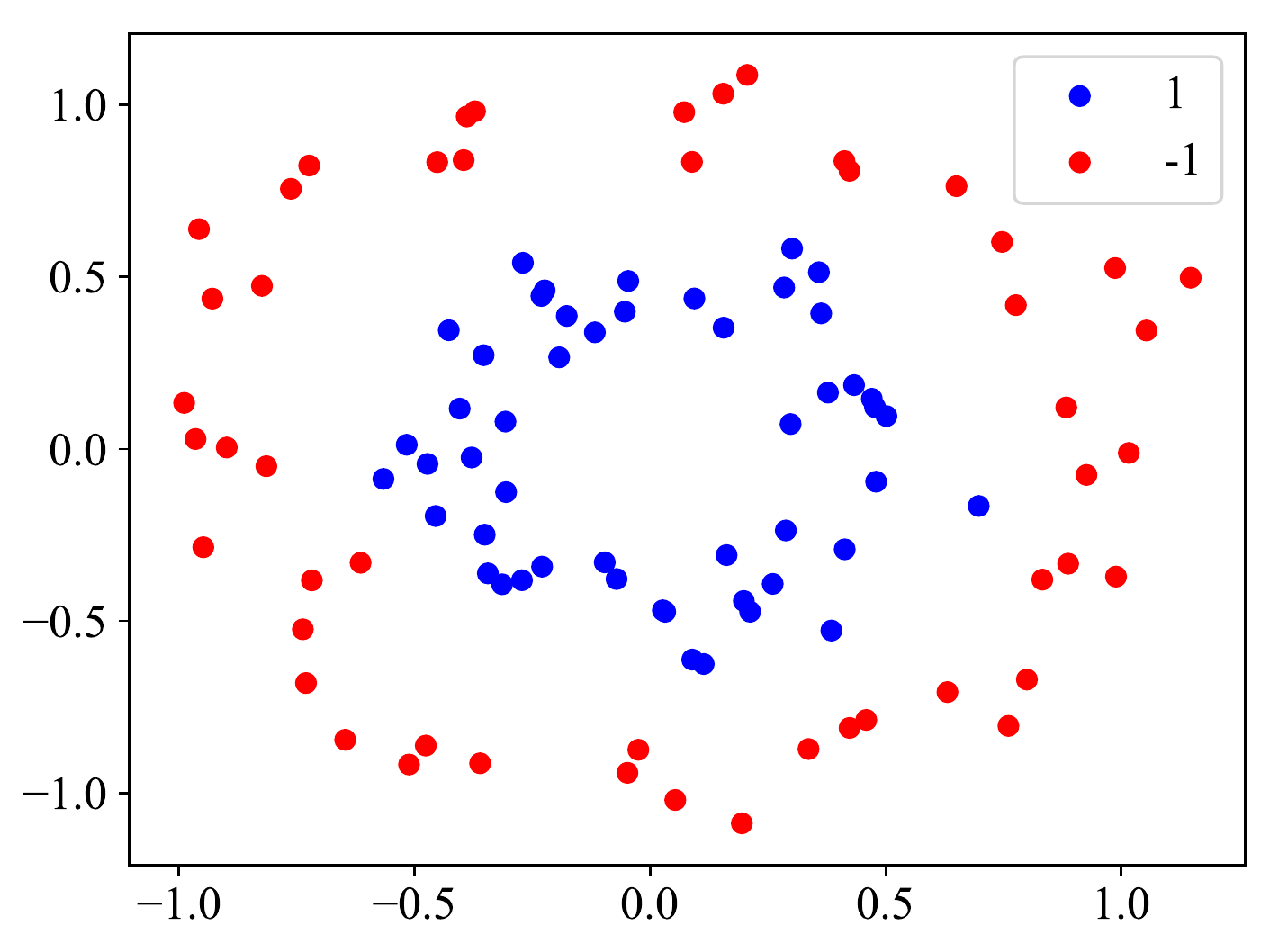}
    \caption{The Circle Data}\label{fig:point_label}
\end{figure}

We here give a show case to explain the interpretation property of \ds~in a direct way. In this case, we generate a circle data set containing samples of two classes, as shown in Fig.~\ref{fig:point_label}. The positive samples are generated from a circle with radius of 0.5 plus a Gaussian noise with variance of 0.1. The negative samples are generated from a circle with radius of 1 plus the same Gaussian noise. We use the circle data set to train a 3-layer \ds~model, where the middle layers contain 40 and 60 base-SVM groups, respectively. Because this experiment is a binary classification, a base-SVM group only contains one base-SVM.

In the experiment, we traverse the feature space from the coordinate point $(-2, -2)$ to $(2, 2)$. For each coordinate point, we calculate the average confidence of all base-SVMs in each layer. Figs.~\ref{fig:heat_h1} - \ref{fig:heat_h3} plot the confidence distribution maps in different layers. The samples with low confidence are thus near to the SVM classification hyperplane, so the low confidence areas in red form the ``classifying plane'' of a layer.

It is obvious that there is a clear ``classifying plane'' generation process from Figs.~\ref{fig:heat_h1} to~\ref{fig:heat_h3}. In the layer 1, the low confidence values concentrate in the center of the map. In the layer 2, the low confidence values have a vague shape as a circle. In the layer 3, the low confidence values distribute as a circle shape that clearly divides the feature space into two parts. This process demonstrates how an \ds~model extracts the data representations of the feature space layer by layer.

\subsection{Parallelization}

The deep stacking network is proposed for parallel parameter learning. For a DSN with $L$ blocks, given a group of training samples, the DSN can resample the data set as $L$ batches. A DSN block only uses one batch to update its parameter. In this way, the training of DSN parameters could be deployed over $L$ processing units. The parallelization of DSN training is in the block level.

This parallel parameter learning property could be further extended by using Parallel Support Vector Machines~\cite{cascade_svm}. Because in a SVM classifier, only the support vector samples are crucial, we could divide a training set as several $M$ sub-sets, and use $M$ virtual SVM classifiers to select support vector candidates from each sub-set. Finally, we use the support vector candidates of all sub-sets to train the final model. In this way, the training of a \ds~block can be deployed over $M$ processing units, and the all \ds~model can be deployed over $L\times M$ processors. That is to say the training of base-SVMs in a block is also parallelizable in \ds. The parallelization of \ds~ training is in the base-SVM level. The parallel degree of whole model is greatly improved.  As reported in Ref~\cite{cascade_svm}, the speed-up for a 5-layer Cascade SVM (16 parallel SVMs) to a single SVM is about 10 times for each pass and 5 times for fully converged.

\section{Experiments}

\subsection{Image Classification Performance}\label{sec:mnist}

We first test the performance of \ds~on the MNIST image classification database~\cite{mnist}. The MNIST database contains 60,000 handwritten digits images in the size of 28 $\times$ 28 for training and validation, and 10,000 images for testing. The images are classified as ten classes according to the digits written on them. The \ds~model used in the experiment consists of three layers -- two middle layers and an output layer. Both of the block in the two middle layers contains 20 base-SVM groups, and each group contains 10 base-SVMs, \emph{i.e.}, 200 base-SVMs one layer. We actually had tried neural networks with deeper layers, but the classification performance could not be improved significantly. The benchmark models includes: $i$) A 3-layer deep stacking network where the block in each middle layer contains 200 hidden neurons; $ii$) As analyzed in the Model Properties section, the \ds~could be considered as a type of neural network. Therefore, we use the BP algorithm to train a same structure \ds~as a benchmark; $iii$) The 3-layer neural network models with the same neuron connection structure as the 3-layer SVM-DSN. In these benchmarks, we use SVM output~\cite{deepsvm}, and the different activate functions in neurons; $iv$) The best 3-layer NN benchmark listed in the homepage of MNIST -- 3-layer NN, 500+300 hidden units, cross entropy loss, weight decay~\cite{mnist}; $v$) A bagging of 41 base-SVM groups, each group contains 10 base-SVMs.

In the \ds~model fine-tuning, we have two hyper-parameters to set, \emph{i.e.}, $C1$ and $C2$ of the base-SVM's objective function. The two hyper-parameters are used to balance structural risks and empirical risks in a base-SVM.
Either too big or too small for the two hyper-parameters may lead model performance degenerate. Therefore, we use the trial and error method to set the hyper-parameters. The learning rate $\eta$ is the other hyper-parameter, which could be dynamic setting using elegant algorithms such as Adam. In our experiment, we directly set the learning rate as a fix value $\eta = 0.0005$ to ensure the experiment fairness.

Table~\ref{tab:fully_connect} gives the MNIST image classification results. The \ds~model achieved the best performance compared with the other benchmarks, which verified the effectiveness of \ds. In the benchmarks, the 3-layer \ds +BP model has the same model structure with \ds~but was trained by the BP algorithm. The results show that \ds~has a better performance than the \ds+BP benchmark, which indicates that the base-SVMs ``do their own best'' feature of \ds~is a positive feature for model performance. In fact, the idea of BLT fine-tuning could be extend to optimize the deep stacking networks with any derivable model as blocks, such as soft decision-making tree and linear discriminant analysis.

\begin{table}\small \centering\caption{MNIST Classification Performance}\label{tab:fully_connect}
\begin{tabular}{l|c}
  \toprule 
  \bf{Models} & \bf{Error Rate (\%)}\\\midrule
  3-layer \ds &  \bf{1.49}\\
  3-layer DSN & 1.65 \\
  3-layer \ds, BP &  {1.62}\\
  3-layer NN, SVM output, sigmoid & 1.74\\
  3-layer NN, SVM output, tanh & 1.59\\
  3-layer NN, SVM output, ReLU & 1.56\\
  Homepage benchmark~\cite{mnist} & {1.53}\\
  Bagging of base-SVMs & 5.41\\
  \bottomrule 
\end{tabular}
\end{table}

\subsection{Feature Extractor Compatibility}

Currently, the mainstream image classifiers usually adopt convolutional neural networks (CNN) as feature extractors. Table~\ref{tab:cnn_minist} demonstrates the MNIST classification performance of \ds~with a CNN feature extractor. In the experiment, we connect a CNN feature extractor with a 3-layer \ds~model. The structure of \ds~is same as in Table~\ref{tab:fully_connect}. The CNN feature extractor contains 3 convolutional layers, and each layer consists of 24 flitters with the 5$\times$5 receptive field. In the CNN and \ds~mixture model, we first pre-trained the CNN part using BP, and then uses the feature extracted by CNN as input of the \ds~part to train the blocks. In the fine-tuning step, the CNN part is fine tuned by BP and the \ds~part is tuned by BLT. The benchmark models include: $i$) The same CNN feature extractor connected with a 3-layer DSN, where the 3-layer DSN has the same structure with the 3-layer \ds~model; $ii$) The CNN+\ds~model trained by the BP algorithm; $iii$) The neural networks consist of 3 CNN layers and 3-layer neural network with different activate functions, where the structures of the CNN and the neural network are same as the CNN + \ds~model; $iv$) The trainable CNN feature extractor + SVMs with affine distortions, which is the best benchmark with the similar models scale listed in the MNIST homepage; $v$) The gcForest with convolutional kernels~\cite{ijcai2017-497}. As shown in Table~\ref{tab:cnn_minist}, the CNN + \ds~model achieved the best performance, which verified the effectiveness of our model again. What's more, this experiment demonstrates that the \ds~model is completely compatible to the neural network framework. The other types of networks, such as RNN and LSTM, could also be used as feature extractors of \ds~to adapt diversified application scenarios.

\begin{table}\small \centering\caption{MNIST Classification with CNN Feature Extractor}\label{tab:cnn_minist}
\begin{tabular}{l|c}
  \toprule 
  \bf{Models} & \bf{Error Rate (\%)}\\\midrule
  CNN  + \ds    &  \bf{0.51}\\
  CNN + DSN    & {0.60} \\
  CNN  + \ds, BP    &  {0.72} \\
  CNN + sigmoid activation    &  0.80\\
  CNN + tanh activation  &  0.67\\
  CNN + ReLU activation   &  0.58\\
  Homepage benchmark~\cite{mnist} &  0.54\\
  gcForest~\cite{ijcai2017-497} & 0.74\\
  \bottomrule 
\end{tabular}
\end{table}

\begin{table}\small \centering\caption{IMDB Classification Performance}\label{tab:imdb}
\begin{tabular}{l|c}
  \toprule 
  \bf{Models} & \bf{Error Rate (\%)}\\\midrule
  \ds    &  \bf{10.51} \\
  DSN    &   11.15   \\
  \ds, BP &  11.42  \\
  Random Forest   & {14.68} \\
  XGBoost    &  {14.77} \\
  AdaBoost   &  16.63  \\
  SVM (linear kernel) & 12.43 \\
  Stacking & 11.55 \\
  Bagging of base-SVMs & 11.66 \\
  gcForest~\cite{ijcai2017-497} & 10.84 \\
  \bottomrule 
\end{tabular}
\end{table}

\subsection{Comparison with Ensemble Models}

In this section, we compare the performance of \ds~with several classical ensemble models. Because on the MNIST data set, the performance of ensemble models are usually not very well. For a fair comparison, we use the IMDB sentiment classification data set~\cite{maas2011learning} in our experiment. Many tree based ensemble methods achieved good performance on this data set~\cite{ijcai2017-497}. The IMDB dataset contains 25,000 movie reviews for training and 25,000 for testing. The movie reviews are represented by tf-idf features and labeled as positives and negatives. The \ds~model used in this experiment consists of 4 middle layers, and the number of base-SVMs in the middle layer are 1024-1024-512-256. The benchmark models include Random Forest, XGBoost, AdaBoost and SVM (linear kernel). The four benchmarks are also stacked as a stacking benchmark~\cite{stacking_kdd}. A bagging of base-SVMs and the grForest~\cite{ijcai2017-497} are also included as competitor. A 4-layer DSN is used as a benchmark, where the number of hidden neurons in the middle layer blocks are same as the number of base-SVMs in the \ds~model. The \ds~model trained by the BP algorithm is also used as the benchmark.

As shown in Table~\ref{tab:imdb}, the \ds~model achieved the best performance again. Especially, the performance of \ds~is better than the stacking benchmark, which indicates that holistic optimized multi-layer stacking of linear base-learners can defeat the traditional two-layer stacking of strong base-learners. In the experiment, the performance of the BLT algorithm is yet better than the BP algorithm. The ``do their best'' feature of base-SVM in BLT is still effective in text sentiment classification.

\section{Related Works}

This work has close relations with SVM, deep learning, and stacking. The support vector machine was first proposed by Vapnik in~\cite{svm}. Multi-layer structures in SVM were usually used as speedup solutions. In cascade SVM~\cite{cascade_svm}, a multi-layer cascade SVM model structure was used to select support vectors in a parallel way. In the literature~\cite{collobert2002parallel}, a parallel mixture stacking structure was proposed to speed up SVM training in the very large scale problems. Before our work, some studies proposed to use SVM to replace the output layer of a neural network~\cite{wiering2013neural,deepsvm}.

In recent years, neural network based deep models has achieved great success in various applications~\cite{deeplearning}. The gcFroest model~\cite{ijcai2017-497} was proposed to use the forest based deep model as an alternative to deep neural networks. The PCANet builds a deep model using unsupervised convolutional principal component analysis~\cite{pcanet}. LDANet is a supervised extension of PCANet, which uses linear discriminant analysis (LDA) to replace the PCA parts of PCANet~\cite{pcanet}. Deep Fisher Networks build deep network through stacking Fisher vector encoding as multi-layers~\cite{dfn}.

The DSN framework adopted in this work is a scalable deep architecture amenable to parallel parameter training, which has been adopted in various applications, such as information retrieval~\cite{dsn}, image classification~\cite{app2}, and speech pattern classification~\cite{dcn}. T-DSN uses tensor blocks to incorporate higher order statistics of the hidden binary features~\cite{tdsn}. The CCNN model extends the DSN framework using convolutional neural networks~\cite{ccnn}. To the best of our knowledge, there are very few works introduce the advantages of SVM into the DSN framework.

Stacking was introduced by Wolpert in~\cite{wolpert1992stacked} as a scheme of combining multiple generalizers. In many real-world applications, the stacking methods were used to integrate strong base-learners as an ensemble model to improve performance~\cite{jahrer2010combining}. In the literature, most of stacking works focused on designing elegant meta-learners and create better base-learners, such as using class probabilities in stacking~\cite{witten}, using a weighted average to combine stacked regression~\cite{rooney2007weighted}, training base-learners using cross-validations~\cite{stacking_kdd}, and applying ant colony optimization to configure base-learners~\cite{chen2014applying}. To the best of our knowledge, there are very few works to study how to optimize multi-layer stacked base-learners as a whole.

\section{Conclusion}

In this paper, we proposed an \ds~model where linear base-SVMs are stacked and trained in a deep stacking network way. In the \ds~model, the good mathematical property of SVMs and the flexible model structure of deep stacking networks are nicely combined in a same framework. The \ds~model has many advantage properties including holistic and local optimization, parallelization and interpretation. The experimental results demonstrated the superiority of the \ds~model to some benchmark methods.

\section{Acknowledgments}
Prof. J. Wang' s work was partially supported by the National Key Research and Development Program of China (No.2016YFC1000307), the National Natural Science Foundation of China (NSFC) (61572059, 61202426), the Science and Technology Project of Beijing (Z181100003518001), and the CETC Union Fund (6141B08080401). Prof. J. Wu was partially supported by the National Natural Science Foundation of China (NSFC) (71531001, 71725002, U1636210, 71471009, 71490723).

\section{Appendix}

{\em {\bf Property:} Given a set of virtual samples $T^{(l,i)} = \{(\bm{x}_{k}^{(l)}, \tilde{{y}}^{(l,i)}_{k})| k = 1, \ldots, K\}$ for $svm(l,i)$, to minimize the loss function defined in Eq.~\eqref{eq:obj_subsvm} is a convex optimization problem.}

{\em Proof.} For the sake of simplicity, we omit the superscripts $(l)$ of $\bm{x}_{k}^{(l)}$ and $\tilde{y}_{k}^{(l)}$ in our proof. We define a constrained optimization problem in the form of
\begin{equation}\label{equ:obj_subsvm_st}
  \begin{aligned}
   \min_{\bm{\omega}, b, \xi_{k}, \hat{\xi}_{k}, \zeta_{k}}  & \frac{1}{2}\left\|\bm{\omega}\right\|^{2} + C_{1}\sum_{k \notin \Theta}\zeta_{k} + C_{2}\sum_{k \in \Theta}\left(\xi_{k} + \hat{\xi}_{k}\right)& \\
    s.t.
    & \quad 1 - \tilde{y}_{k}\left(\bm{\omega}^\top \bm{x}_{k} + b\right) \leq \zeta_{k}, & (1) \\
    & \quad \zeta_{k} \geq 0, \; k \in \Theta;\\
    & \quad \left(\bm{\omega}^\top \bm{x}_{k} + b\right) - \tilde{y}_{k} \leq \epsilon + \xi_{k}, & (2) \\
    & \quad \tilde{y}_{k} - \left(\bm{\omega}^\top \bm{x}_{k} + b\right) \leq \epsilon + \hat{\xi}_{k}, & (3) \\
    & \quad \xi_{k} \geq 0, \; \hat{\xi}_{k} \geq 0, \; k \notin \Theta.\\
  \end{aligned}
\end{equation}
We can see the constrained optimization problem Eq.~\eqref{equ:obj_subsvm_st} is in a quadratic programming form as
\begin{equation}\label{}
  \begin{aligned}
    \min_{\mathbf{a}} &\quad \frac{1}{2}\mathbf{a}^{\top}\mathbf{U}\mathbf{a} + \mathbf{c}^{\top}\mathbf{a} \\
    s.t. & \quad \mathbf{Q}\mathbf{a} \leq \mathbf{p},
  \end{aligned}
\end{equation}
where $\mathbf{a} = ({\bm{\omega}, b, \bm{\xi}, \bm{\hat{\xi}}, \bm{\zeta}})$, and $\mathbf{U}$ is a positive semi-definite diagonal matrix. Therefore, the constrained optimization problem is a quadratic convex optimization problem~\cite{convex_optimization}. It is easy to prove that the constrained optimization problem defined in Eq.~\eqref{equ:obj_subsvm_st} is equivalent to the unconstrained optimization problem defined in Eq.~\eqref{eq:obj_subsvm}~\cite{zhang2003statistical}.
Therefore, the optimization problem of base-SVM is equivalent to the problem defined in Eq.~\eqref{equ:obj_subsvm_st}. The optimization problem of base-SVM is a convex optimization problem.

\bibliographystyle{aaai}

\end{document}